\def\year{2021}\relax
\documentclass[letterpaper]{article} 
\usepackage{aaai21}  
\usepackage{times}  
\usepackage{helvet} 
\usepackage{courier}  
\usepackage[hyphens]{url}  
\usepackage{graphicx} 
\usepackage{natbib}  
\usepackage{caption} 
\usepackage{algorithmicx}
\usepackage{amsmath}
\usepackage{amsfonts}
\usepackage{amssymb}
\usepackage{algorithm}
\usepackage{algpseudocode}
\usepackage{mathtools, nccmath}
\usepackage{float}
\usepackage{subfig}
\usepackage{graphicx}
\usepackage{xcolor}
\usepackage{multicol}
\usepackage{caption}
\DeclareMathOperator*{\argmax}{arg\,max}
\newcommand{\rb}{\textsl{RB2 }}
\newcommand{\f}{\ensuremath{\mathcal{F}}}

\title{Relational Boosted Bandits}
\author {

        Ashutosh Kakadiya,\textsuperscript{\rm 1}
        Sriraam Natarajan, \textsuperscript{\rm 2}
        Balaraman Ravindran \textsuperscript{\rm 1} \\
}
\affiliations {
    \textsuperscript{\rm 1} Robert Bosch Centre for Data Science and Artificial Intelligence, Indian Institute of Technology Madras\\
    \textsuperscript{\rm 2} The University of Texas at Dallas\\
    
    kashutosh@cse.iitm.ac.in, 
    Sriraam.Natarajan@utdallas.edu, 
    ravi@cse.iitm.ac.in
}

\begin{document}

\maketitle

\begin{abstract}
Contextual bandits algorithms have become essential in real-world user interaction problems in recent years. However, these algorithms represent context as attribute value representation, which makes them infeasible for real-world domains like social networks, which are inherently relational. 
We 
propose {\it Relational Boosted Bandits} (\textsl{RB2}), a contextual bandits algorithm for relational domains based on (relational) boosted trees. \rb enables us to learn interpretable and explainable models due to the more descriptive nature of the relational representation. We empirically demonstrate the effectiveness and interpretability of \rb on tasks such as link prediction, relational classification, and recommendation.
\end{abstract}
\section{Introduction}
The contextual bandit framework has gained a lot of attention in several real-world personalization applications from news recommendation to online advertising~\cite{L.Li}, comments recommendation~\cite{D.Mahajan}, clinical trials~\cite{A.Durand}, A/B testing, and dialogue systems ~\cite{B.Liu}. Contextual bandit is an extension of {\em multi-armed bandits} with a context vector for each user. This context about the individual user allows the personalization of the actions than calculating a simple argmax over all actions. The general framework of contextual bandits~\cite{introcb} can be formalized as follows: at each time instance $t$, a user arrives with a vector of information (or features) referred to as a context vector. The goal is to choose one action for the user among $K$ actions given the context (user and actions). The reward $r$ is observed for only the chosen action and the objective is to maximize the total reward over time.

\par Most of the contextual bandit algorithms~\cite{L.Zhou} have focused on propositional domains, where the context is described using a flat feature-vector representation. Typically however, many real-world domains are naturally structured and are described by interacting objects and relations between them. This representation allows for learning richer models. Additional domain knowledge in the form of inductive/search bias is typically employed for learning in such domains. Inspired by this direction, we explore the combination of contextual bandit and Statistical Relational Learning (SRL).   


\par SRL~\cite{book_srl} combines the power of relational/symbolic representations with the ability of probabilistic/machine learning models to handle uncertainty. Consequently, it is well suited for several real-world tasks such as social network analysis, recommendation and biomedical applications. Initial research focused on the development of several models -- directed models~\cite{prm,BLP}, undirected models~\cite{MLN} and bi-directed models~\cite{neville07}. More recently, focus has rightly turned to learning SRL models~\cite{book_srl,problog,RRTGB}. Arguably, learning in these models is computationally intensive as it requires exploring multiple levels of abstractions (at the object level, partial instance level or fully ground level). One of the recent successful methods is to learn a set of relational regression trees using gradient-boosting~(RRTGB)~\cite{RRTGB}. The key advantage of this method is that it learns a set of weak classifiers and can avoid searching for a single model. This method has been successfully applied for many applications such as recommendation~\cite{yang17}, cardiovascular conditions~\cite{natarajan13}, PPMI~\cite{dhami17} and rare diseases~\cite{macleod16} to name a few. While successful, these models are restricted to only batch mode learning. While a previous online algorithm exists~\cite{huynh11} for learning an undirected model, this algorithm does not have the distinct advantage of bandit approaches - the ability to explore or exploit. 

Motivated by the success of the boosting method in batch settings, we propose a new {\em relational contextual bandit framework} based on RRTGB. The resulting framework, called {\em Relational Boosted Bandits} (\textsl{RB2}), combines the power of a powerful learning algorithm with the exploration-exploitation abilities of bandits. One of the key motivations of this combination is that the structure and parameters of the underlying model can be learned simultaneously, thus allowing for effective learning in online settings. Our {\bf key insight} is to represent the policy as a set of relational regression trees (RRT)~\cite{tilde}. Consequently, the online relational learning algorithm \rb has the ability to learn using relational data while employing an effective sampling mechanism to handle the exploration-exploitation trade-off. In addition, since these trees are essentially combined using a simple addition operator, we construct a final tree based on these different trees to build an interpretable model.


\par We make the following key contributions: (1) To the best of our knowledge, we are proposing the first contextual bandit algorithm based on SRL for relational domains.  (2) We propose a parameter-free sampling algorithm for the online learning of probabilistic relational models. (3) Finally, we perform comprehensive experiments on several tasks and demonstrate the efficiency and effectiveness of the proposed approach.

\par The rest of the paper is organized as follows. After introducing the necessary background and related work about contextual bandit and boosted relational regression trees, we propose our \rb algorithm. Then we demonstrate an empirical assessment of our algorithm on real-world data sets before concluding by outlining the areas of future research.

\section{Background and Related work}
\subsection{Contextual Bandit}
Contextual Bandit is a variant of the classical multi-armed bandit problem~\cite{introcb} where the agent has access to side information (context) for better decision making. The agent has to make context-based sequential decisions from time $t= 1, 2, 3, ..., T$. At each time-step, the agent has to decide which arm to select from the given $K$ arms. After selecting the arm, the agent receives the payoff from the environment corresponding only to the selected arm while the other payoffs are unknown. The goal is to learn the policy that maximizes the expected payoff. The arm with the highest expected payoff may be different for different contexts. The context includes static and dynamic information about both the agent and the arms. The typical evaluation metric used is the {\em regret} which is given by the cumulative sum of the difference between optimal payoff and the actual payoff received over the horizon $T$. Many real-world problems can be formulated in a contextual bandit setting~\cite{appli_cb}. Few examples are recommendation systems, financial portfolio management, ad placement on websites, and healthcare. While successful, they are not generally applied to multi-relational settings, a key direction in our work.
\subsection{Boosted Relational Regression Trees}
Gradient-boosted relational regression trees (RRTGB)~\cite{RRTGB,imitation_RRT} adapts gradient-boosting (GB)~\cite{friedman2001} to relational domains. For classification, typically GB, calculates the functional-gradient of the examples in the form $(x_i,y_i)$,  $i=1,2,3,..,M$ and $y_i \in \{1,2,3,..,K\}$. This gradient is the difference between the true label (represented by an indicator function) and the predicted probability of the true label. RRTGB uses a relational regression tree learner (TILDE)~\cite{tilde} to represent the potential function $\psi$. In a relational regression tree, each node represents the conjunction of literals.  In RRTGB, the functional gradient ascent starts with the initial potential function $\psi_0$, iteratively computes gradients and adds to the existing model. Formally at the end of $n^{th}$ iteration, the potential function is given as,
\begin{equation}
    \psi_n = \psi_0 + \Delta_1 + \Delta_2 + ... + \Delta_{n} 
\end{equation}
And, the functional gradient $\Delta_n$ at iteration $n$ is given as,
\begin{equation}
    \Delta_{n} = \eta_n \times E_{x,y}[\partial / \partial \psi_{n-1} log P(y|x;\psi_{n-1}) ]
\end{equation}
This procedure is repeated until a fixed number of iterations is reached or till convergence. Our algorithm uses RRTGB as a base learner to model the relation between context and the probabilities of getting a reward of $1$.

\section{Problem Formulation}
For every time-step, $t= 1, 2, 3,..., T$, a user arrives with context as $x_t$. 
The features consist of the defined predicates of both users and the arms. At a fairly high level, \rb will choose the action $a_t \in \{1,2,3,..,K\}$, and obtains the reward $r_{t,a_t} \in \{0,1\}$ associated with the chosen arm $a_t$ and context $x_t$. The goal is to maximize the (expected) cumulative sum of reward.

Let $p(x,a)$ denote the probability of observing a reward $r = 1$. The model predicts $p(x,a)$ $\forall a$ and selects the arm $a_t = \argmax_a p(x,a)$. The regret is simply the difference between reward received by the algorithm $r_{t,a_t}$ and reward $r_{t,a_t*}$ associated with the optimal action $a*$. Formally, at time period $T$, the cumulative regret $R(T)$ is defined as,
\begin{equation}
    R(T) = \sum_{t=1}^{T}  (r_{t,a^*_t} -r_{t,a_t})
\end{equation}
We use cumulative regret as the evaluation metric. 
\section{Relational Boosted bandits}
A key aspect of our setting is that the context is inherently relational, i.e., the context cannot be specified with a flat feature vector based representation. Instead, each instance's attributes/relations could be of differing size (papers published, movies acted, hospital visits, lab tests, etc.) thus requiring a representation that is more general than a simple feature vector. To this effect, we employ first-order logic based notations for representation and learning. A (logical) \textbf{predicate} is of the form $\mathcal{R}(t_1, \dots, t_k)$ where $\mathcal{R}$ is a predicate and $t_i$ are {\bf arguments} or logical variables. We refer to the totality of observed contexts as {\em background knowledge}. A \textbf{substitution} is of the form $\theta = \{\langle v_1, \dots, v_k \rangle/ \langle t_1, \dots, t_k\rangle\}$ where $v_i$s are logical variables and $t_i$s are terms. A \textbf{grounding} of a predicate with variables $v_1, \dots, v_k$ is a substitution $\{\langle v_1, \dots, v_k \rangle \slash \langle V_1, \dots, V_k \rangle\}$\footnote{We use uppercase for predicates/groundings and lowercase for variables.} mapping each of its variables to a constant in the domain of that variable. 

\noindent \fbox{
 \parbox{0.98\columnwidth}{
\textbf{Given:} The context $x_t$  at time $t$ and the accumulated background knowledge $\mathcal{B}$  \\
\textbf{To Do:} Predict $a_t =  \mathcal{F}(P(x_t, a_t |\mathcal{B}))$
}}
The goal of our system is to pick an action based on the learned action model $P$ and the exploration-exploitation strategy $\f$. To learn $P$, we employ gradient-boosting while for $\f$, we consider two different types of strategies -- epsilon-greedy and informed sampling. We now present the details of the formulation and the learning methodology. 

\subsection{\rb framework}
The key aspect of our framework is that it is an {\bf online} algorithm. The only related prior work that considered online learning of MLNs is by Huynh and Mooney (year{huynh11}). Our approach can be seen as a more efficient approach that explicitly learns to the trade-off between exploration and exploitation. More precisely, \rb is an {\bf online} algorithm where the learning happens in mini-batches. The exploration strategy is employed in the choice of batches while training and the action selection during evaluation. During training, the action choices are aggregated into mini batches and the parameters are updated once every mini batch. During deployment/testing, there is no aggregation since there is no learning. We now proceed to explain each of the components in greater detail. We focus on learning the distribution before discussing the action selection.

At each training mini-batch $b$, a certain set of training examples have been collected. From these examples, a certain set of examples (say $D_b$) will be selected based on the exploration strategy that we will explain next. For each example, given its context $x_t$, the currently accumulated background knowledge $\mathcal{B}$, the goal is to learn $P(a_t,x_t,\mathcal{B})$ which gives the probability distribution over the actions given the context. In our bandit setting, this corresponds to the choice of multiple arms. In a classification setting, this could correspond to the distribution over the classes, in a recommendation setting over the items to be recommended etc. 

As mentioned earlier, we model the distribution $P(a_t,x_t,\mathcal{B})$ using RRTGB. Hence the distribution $P(a_t,x_t,\mathcal{B})$ is represented as, 
\begin{equation*}
    P(a_t,x_t,\mathcal{B}) = \frac{e^{\psi(a_t,x_t,\mathcal{B})}}{1+e^{\psi(a_t,x_t,\mathcal{B})}}
\end{equation*}
RRTGB now learns the gradient of the loglikelihood over the mini-batch $\sum_t[log( P(a_t,x_t,\mathcal{B}))]$ w.r.t $\psi$. Following the standard gradient-boosting, point-wise gradients are computed for each example. This is simply of the form $I(a=a_t)-P(a_t,x_t,\mathcal{B})$ which is the difference between whether the action was chosen in the given data and the current predicted probability of the action given the context and background knowledge. These point-wise gradients are chosen for all the examples in $D_b$. Next a TILDE tree is fit over these gradients, where the goal of the tree learner is to minimize the weighted variance of the regression value. Once the trees are fitted, boosting proceeds to the next iteration where newer gradients are computed and the new tree is fit. The process is repeated for the preset number of boosting iteration in each minibatch. Typically, in our experiments, the number of trees (K) is preset to $4\leq K\leq 10 $.

We update the background knowledge $\mathcal{B}$ periodically as new relational information arrives. Typically, this is in the form of newer attributes of either known objects/entities or new objects themselves. For example, in a university domain, this could be newer courses offered by the department or in a movie domain, this could be newer movies directed/acted by a particular person. This new information could also be a modified background knowledge, for instance, new merger/acquisitions of the concerned firms. One of the advantages of our approach is that we can {\bf adapt to the changing background knowledge} more seamlessly because the later iterations of the boosted model can appropriately model the target distribution to better reflect this updated knowledge.

Given that we have explained how $P(a_t,x_t,\mathcal{B})$ is learned, we now turn our focus on effective sampling strategy for online learning. We note that this sampling algorithm is specifically useful in relational domains. 
The goal of this sampling algorithm is to assign a high probability of getting sampled to important samples. If we use uniform sampling then important data samples have very less chance to getting selected. We use stochastic prioritization to assign priority(weights) to data samples. To achieve this, we divide each mini-batch data set $\mathcal{D}$ into two data sets $\mathcal{D}_p$ and $\mathcal{D}_n$. $\mathcal{D}_p$ consists of all data samples with $r=1$ and $\mathcal{D}_n$ with $r=0$. We can easily obtain the priority probability distribution $P_s(i)$ for data sample $i \in \mathcal{D}_p$ by Equation \ref{dp} 
\begin{equation} \label{dp}
    P_s(i) = \frac{e^{1-p_{i}}}{\sum_{j\in\mathcal{D}_p}e^{1-p_{j}}}
\end{equation}
and for data sample $i \in \mathcal{D}_n $ by equation \ref{dn}.
\begin{equation} \label{dn}
    P_s(i) = \frac{e^{p_{i}}}{\sum_{j\in\mathcal{D}_n}e^{p_{j}}}
\end{equation}



Recall that RRTGB predicts the probability $p$ of choosing the correct action, i.e., $p(a_t) = 1 \forall t$. This corresponds to getting a reward $1$. As with classification, the goal is to iteratively make $p \to 1$ for positive samples. 
To improve the prediction, our model should predict with higher $p$ for $i \in \mathcal{D}_p $ and lower $p$ for $i \in \mathcal{D}_n $ after each batch training. This necessitates assignment of higher prirority to samples $i \in \mathcal{D}_p $ with lower $p$ and samples $i \in \mathcal{D}_n $ with higher $p$.
 Intuitively, lower confidence indicates that a model has not learned about these samples. This is achieved by employing the sampling probabilities given by equation \ref{dp} for data sample $i \in \mathcal{D}_p $ and equation \ref{dn} for data sample $i \in \mathcal{D}_n $. During each batch update, the goal is to obtain a batch of samples with this distribution and train the model incrementally on it. This stochastic prioritization will also better help to avoid overfitting the model than simply sampling greedily. While we demonstrate this aspect empirically, it can be easily understood by observing that our model simply does not pick up the top k most uncertain samples in the batch but samples based on a priority distribution. We now formalize the algorithm.
 
\subsection{Algorithm}
Algorithm \ref{rbsmain} outlines the Relational Boosted Bandits procedure. Let us denote the data set buffer as $\mathcal{D}.$ Data set $\mathcal{D}_l$ is used to train the boosting classifier and consists of the set of tuples represented as $(x,a,r)$. Initially, data set $\mathcal{D}_l$ is constructed using any random policy. This is due to the lack of an informed prior on the policies. If there are informed priors as inductive biases, domain knowledge and/or constraints on the samples, they could be incorporated easily. 

We learn the first set of $K$ relational boosted trees and this is denoted by $\mathcal{F}_0$. At every time stamp $t$, a user  arrives with context $x_t$ and we sample action $a_t$(line $\textbf{6}$) according to softmax distribution,
\begin{equation*}
    P(a_i) = \frac{e^{(p( a_i/x_t,\mathcal{B} )) /\tau }}{\sum_{j=1}^{K}e^{(p( a_j/x_t,\mathcal{B} )) /\tau }}   
\end{equation*}
In the softmax, $\tau$ controls the degree of exploration, i.e., when $\tau = 0$, the arm is chosen purely greedy. When $\tau \rightarrow \infty $, the algorithm selects randomly. Reward $r_t\in \{0,1\}$ is then obtained for selected arm $a_t$. Line \textbf{8} shows that data set $\mathcal{D}_i$ is updated with new data sample $(x_t,a_t,r_t,p_t)$. For the periodical batch-update, we sample the data set $\mathcal{D'_i}$ according to Algorithm \ref{infsamp}. In line \textbf{12} we incrementally update the model $\mathcal{F}_i$ by adding a new set of $K$ relational trees to the already fitted model $\mathcal{F}_{i-1}$. These new trees are trained on sampled data set $\mathcal{D'}$. In general, at $i^{th}$ batch update, new $K$ trees will be added to previous $(i-1)*K$ trees. This process is repeated as each batch of the data arrives. The final set of trees is then returned.

\begin{algorithm}[]
\caption{Boosted Relational Bandits: Softmax Exploration with Informed Sampling} \label{rbsmain}
\begin{algorithmic}[1]
\State Define $\mathcal{D} = \mathcal{D}_{l} $
\Comment{Logged data, gathered by arbitrary policy}
\State $\mathcal{F}_{0} \coloneqq \text{RRTGB} (\mathcal{D}_{l}) $  
\Comment Cold start training
\For {batch i = 1,2,...,N }
    \For{t = 1,2,...,T}
    
        \State Observe context $x_t$
        \State Sample $a_t \sim \mathcal{F}( p( a/x_t,\mathcal{B} ) ) $
        \Comment Softmax Action selection under learned distribution. 
        \State Receive reward $r_{t} \in \{0,1\}$ 
        \Comment Obtain 0/1 reward
    
        \State $\mathcal{D}_i = \mathcal{D}_i \cup \{(x_{t},a_{t},r_{t},p_{t})\} $
        \Comment Update $\mathcal{D}_i$
        \State Update the background knowledge $\mathcal{B}$
    \EndFor
    \State $\mathcal{D'}_i \sim \text{Informed Sampling}(\mathcal{D}_i) $
    \Comment{Sample a batch of data using Algorithm\ref{infsamp}}
    \State $\mathcal{F}_{i} \coloneqq \mathcal{F}_{i-1} + \text{RRTGB}(\mathcal{D'}_i)$ \Comment Update model by adding new $K$ trees learned on $\mathcal{D'}_i$

\EndFor
\end{algorithmic}
\end{algorithm}
\begin{algorithm}[]
\caption{Informed Sampling: Stochastic Prioritization,} \label{infsamp}
\begin{algorithmic}[1]
\Function{InformedSampling}{$\mathcal{D}$}
\State $\mathcal{D}_{p} \coloneqq \mathcal{D}[r = 1]$
\Comment{Examples with reward 1}
\State $\mathcal{D}_{n} \coloneqq \mathcal{D}[r = 0]$
\Comment{Examples with reward 0}    
\For{k=1 to K/2} \Comment{K is batchsize}
    \State Sample transition $i $ from $\mathcal{D}_{p} $ $ \sim  P_s(i) = \frac{e^{1-p_{i}}}{\sum_{j}e^{1-p_{j}}} $
    \Comment Sample transition from positive examples
    \State $\mathcal{\tilde{D}}_{p} \coloneqq \mathcal       
            {D}_{p} \cup i $ 
    \State $\mathcal{D}_{p} \coloneqq \mathcal{D}_{p}              \setminus \{i\} $       
    \Comment Remove for avoiding re-sampling of grounding examples
    \State Sample transition $k $ from $\mathcal{D}_{n} $ $ \sim            P_s(k) = \frac{e^{p_{k}}}{\sum_{j}e^{p_{j}}} $
    \Comment Sample transition from negative examples.
    \State $\mathcal{\tilde{D}}_{n} \coloneqq              \mathcal{\tilde{D}}_{n} \cup k $    
    \State $\mathcal{D}_{n} \coloneqq \mathcal       
            {D}_{n} \setminus \{k\} $       
    \Comment Remove for avoiding re-sampling of grounding examples
\EndFor
\State $\mathcal{\tilde{D}} \coloneqq \mathcal{\tilde{D}}_{p} \cup \mathcal{\tilde{D}}_{n} $
\Comment Merge sampled data sets
\State \Return $\mathcal{\tilde{D}}$
\EndFunction

\end{algorithmic}
\end{algorithm}
Algorithm \ref{infsamp} presents the sampling algorithm for online learning of gradient boosted RRTs. Note that this is another {\em important contribution of our work.} As far as we are aware, apart from the work of Huynh and Mooney~\shortcite{huynh11}, this is one of the few online algorithms for SRL models and the first online algorithm for gradient boosting RRTs.

Lines \textbf{2-3} refer to the division of data set into $\mathcal{D}_p$ and $\mathcal{D}_n$. We sample the data using stochastic prioritization using  Equation~\ref{dp} for $i \in \mathcal{D}_p$ and Equation~\ref{dn} for $i \in \mathcal{D}_n$. Lines 7 and 10 demonstrate the removal of the sampled data sample to avoid re-sampling. If a batch contains multiple occurrences of the same ground atom, the algorithm will consider these several groundings as a single data sample at training time. The data sample is removed after getting sampled from the data set to mitigate this problem. At the end, both data sets are merged into a new data set $\mathcal{\tilde{D}}$ where incremental training is performed.

\subsection{Explainability of the model}

A natural question arises about the interpretability and explainability of the learned model. This is particularly true because our underlying model is based on boosting. A key advantage of our model is that this underlying combination functions of these trees is a {\bf sum}. Hence, these trees could be combined analytically similar to how First-order decision diagrams could be added~\cite{joshi07}. The key difference is that unlike the more general relational structures, our trees have single path semantics, which makes the combination more cumbersome. For instance, multiple instances of the same object in different trees require unification and this needs to be performed repeatedly in domains with a small number of predicates. Finally, one could also approximate these sums by removing branches where the differences in regression values are under a predefined threshold $\delta$.

We take a more empirical approach suggested by Craven and Shavlik~\shortcite{craven95} who proposed constructing tree structured representations of neural nets. The high-level idea is to train a neural network and then make predictions on the training data. Given these predictions, one could simply learn a tree structured model on these relabeled data (based on the original model). We take a similar approach. After the boosted model is trained, we make predictions on the entire data set and then use the predictions to train a single relational regression tree (TILDE tree). This tree has the distinct advantage of being explainable. We demonstrate such trees learned in our domains. We now discuss our experiments. 

\section{Empirical Evaluation}
Our experimental evaluation explicitly aims to answer the following specific questions:
\begin{enumerate}
\item {\bf Effectiveness:} How does \textsl{RB2} perform against other baselines on real-world data?
\item {\bf Necessity of a complex model:} How does \textsl{RB2} perform against linear approximation model?
\item {\bf Interpretability:} Can the resulting model of \textsl{RB2} be interpretable? 
\end{enumerate}
   
\subsection{Data sets}
We assessed the empirical performance of our model on a synthetic Movie Recommendation data set and five real-world relational data sets (Table\ref{table1}). For the simulated movie data set, we created predicates and relations such as {\em user information, good movie, friends, similar movies} etc. The goal in this domain is to predict on the user clicking a suggested movie. For this work, we created about 80k facts to allow for testing the scale of the learning model. We now explain the real data sets. Movie Lense data set~\cite{rdata_repo} contains predicates that cover the relations such as {\em user age, movietype, movie rating} etc. The goal is to predict genres of the movies.  Drug-Drug Interaction(DDI)~\cite{DDI_dataset} has information like {\em Enzyme, Transporter, EnzymeInducer} etc. The goal is to predict the interaction of two drugs. ICML Co-author~\cite{ICML_dataset} includes {\em affiliation, research interests,location} etc. The goal is to predict if two persons worked together in a paper. IMDB~\cite{imdb_data} contains predicates and relations such as {\em Gender, Genre, Movie, Director} etc. We predict the target {\em WorkUnder}, i.e., predict if an actor works under a director. We also employ the sports data of Never Ending Language Learner (NELL) data set~\cite{NELL_dataset} that includes {\em information of players, sports, league information}. The goal is to predict which specific sport does a particular team plays. 
\begin{table}[h]
\centering

\begin{tabular}{|c|c|c|}
\hline
\textbf{data set}      & \textbf{\begin{tabular}[c]{@{}c@{}}Number \\ of Facts\end{tabular}} & \textbf{Target} \\ \hline
Simulated Movie  & 85565& willclick       \\ \hline
IMDB                  & 938& Workunder      \\ \hline
Movie Lense           & 166486& like            \\ \hline
ICML Co-Author        & 1395& CoAuthor        \\ \hline
Drug Interaction   & 1774& interaction     \\ \hline
Sports NELL                &7824 & teamplayssport  \\ \hline
\end{tabular}
\caption{Description of data sets that used in experiments.}
\label{table1}
\end{table}     
\begin{table}[h]
\centering
\begin{tabular}{|c|c|c|}
\hline
\textbf{data set}      & \textbf{\begin{tabular}[c]{@{}c@{}}Batch \\ Size\end{tabular}} & \textbf{\begin{tabular}[c]{@{}c@{}}Trees per \\ batch (K)\end{tabular}} \\ \hline
Simulated Movie Lense & 256  & 8 \\ \hline
IMDB                  & 128 & 6\\ \hline
Movie Lense           & 256 & 5\\ \hline
ICML Co-Author        & 128  & 8\\ \hline
Drug Interaction   & 128& 5\\ \hline
Sports NELL                  & 128& 6\\ \hline
\end{tabular}
\caption{Hyperperameters used in experiments}
\label{table2}
\end{table}
          \newcommand{\imagesize}{5.5cm}
        \begin{figure*}[h]
        \centering
        \subfloat[Synthetic Movie data set ] {\includegraphics[width=\imagesize]{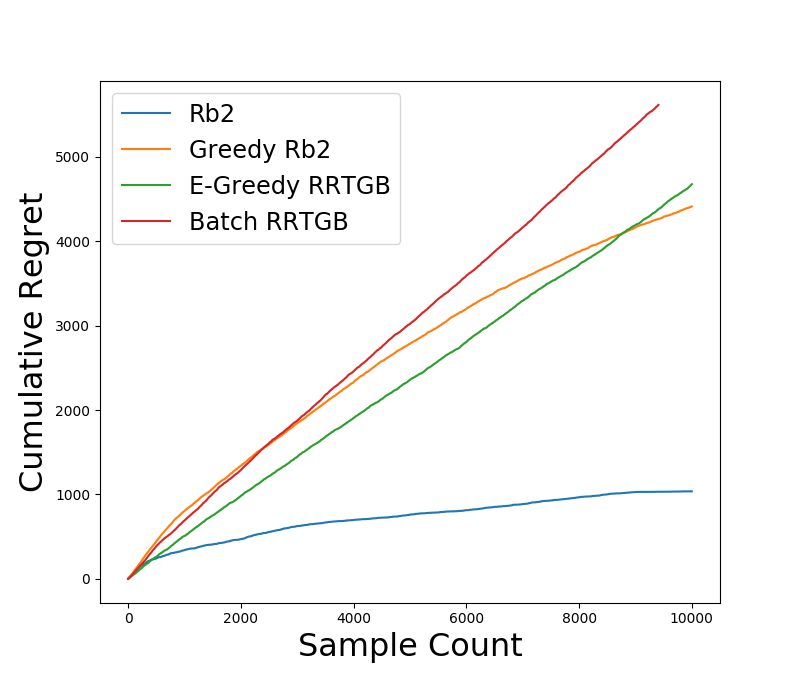}}\hfil
        \subfloat[IMDB Datset]{\includegraphics[width=\imagesize]{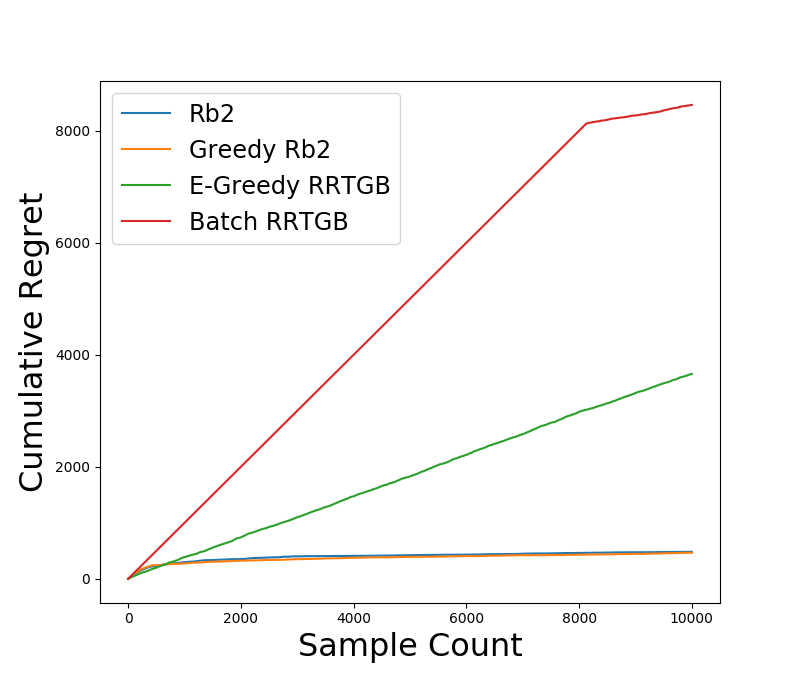}}\hfil
        \subfloat[Drug Interaction data set] {\includegraphics[width=\imagesize]{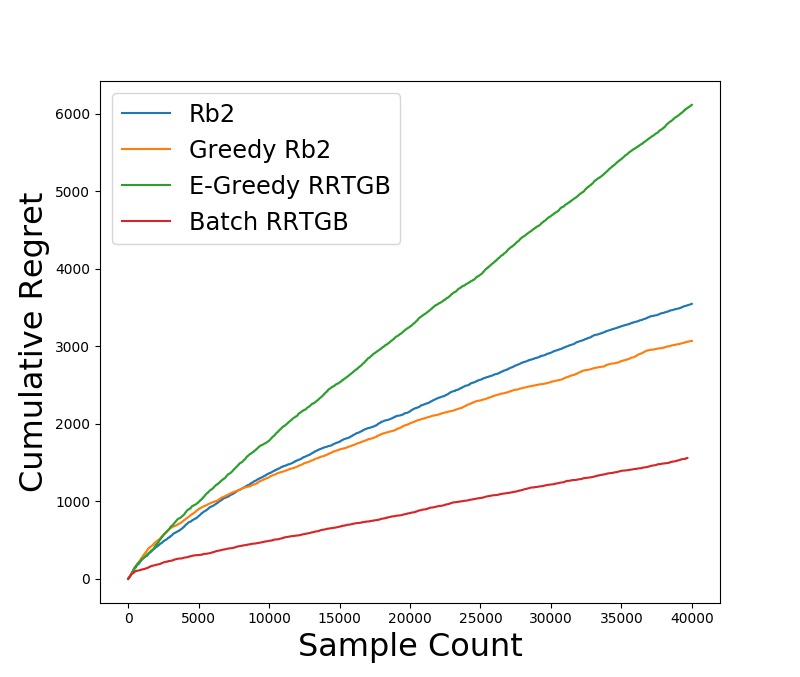}} 
        
        \subfloat[Movie Lense data set] {\includegraphics[width=\imagesize]{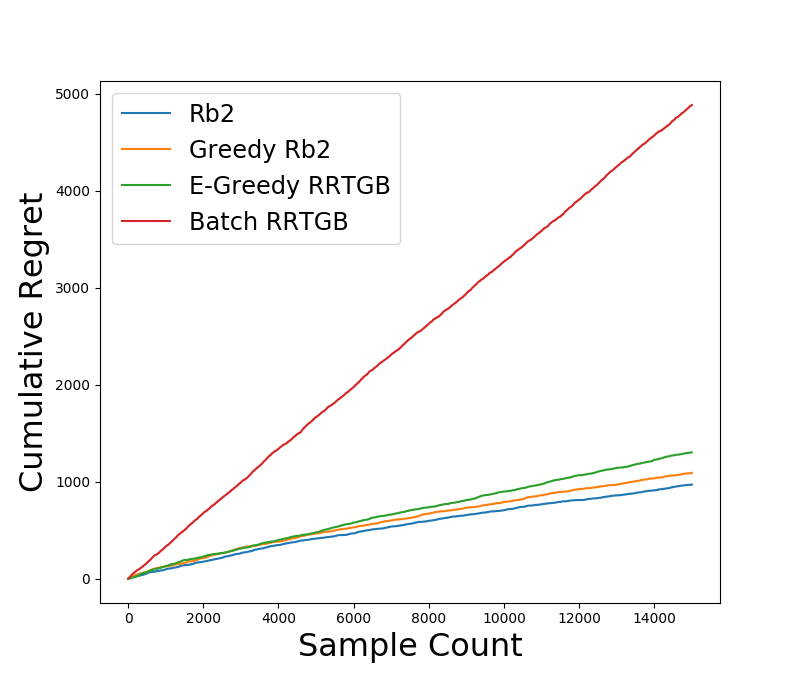}}\hfil   
        \subfloat[ICML Co-Author data set] {\includegraphics[width=\imagesize]{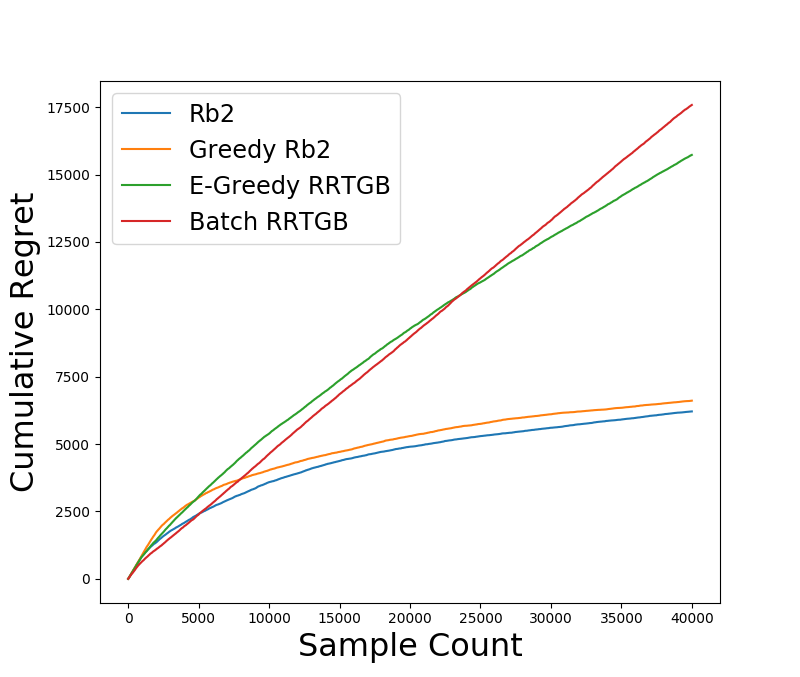}}\hfil
        \subfloat[NELL-Sports data set ] {\includegraphics[width=\imagesize]{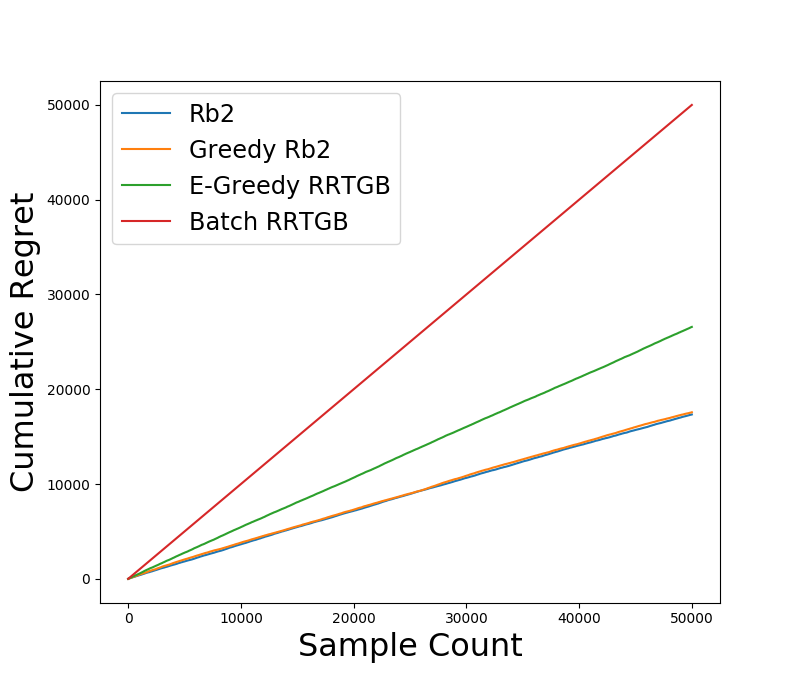}}
        \caption{Comparison of cumulative sum of regret on various relational data sets. Regret of 0 iff algorithm predicts correct labels for a given data point, otherwise 1.}\label{results}
        \end{figure*}
\subsection{Benchmark Algorithms}
    We compared our \rb algorithm with following algorithms:
    \begin{enumerate}
        \item Batch RRTGB (No exploration): We consider an online SRL learning variant of RRTGB~{RRTGB} as a baseline. The model learns a set of trees incrementally on a batch of data, as explained in the algorithm section without any exploration. We used random sampling to sample a batch of data for batch training. This will allow to establish if {\bf exploration indeed helps} in learning a better model.
        \item Epsilon Greedy RRTGB: $\epsilon-$greedy is the standard baseline for multi-armed bandits. We use batch RRTGB as a base learner with $\epsilon-$greedy exploration. Action selection will be made as described in equation \ref{e_greedy}. With probability $\epsilon$, we explore uniformly selected random action among all $K$ actions. Else we exploit the best action among all $K$ actions. This is a {\bf relational epsilon greedy baseline}.
        \begin{equation} \label{e_greedy}
            a \coloneqq \left\{\begin{matrix}
            {\arg\max }_{a} \ \mathcal{ F}_{t-1}(\text{target}(x_{t},a)) &   \text{with } 1-\epsilon  \\
            \text{a uniformly random action}  &  \text{with }  \epsilon
            \end{matrix}\right. 
        \end{equation}
        \item Greedy \textsl{RB2}:  This is a variant of the proposed \rb algorithm. To demonstrate the {\bf effectiveness of informed sampling} described in algorithm \ref{infsamp}, we compared it with greedy sampling described in algorithm \ref{greedysamp}. In greedy sampling, we act greedily by picking the best $K$ data points by sorting them on probability $p_i$.   
        \begin{algorithm}[H]
        \caption{Greedy Sampling} \label{greedysamp}
        \begin{algorithmic}[1]
        \Function{GreedySampling}{$\mathcal{D}$}
        \State $\mathcal{D}_{p} \coloneqq \mathcal{D}[r = 1]$
        \Comment{Set of examples with reward 1}
        \State $\mathcal{D}_{n} \coloneqq \mathcal{D}[r = 0]$
        \Comment{Set of examples with reward 0}    
        \State Sort the data set $\mathcal{D}_{n}$ according to ${1-p_i}'s$

        \State Sort the data set $\mathcal{D}_{p}$ according to $p_i's$

        \State $\mathcal{\tilde{D}}_{p} \coloneqq \mathcal{D}_{p}[0:K/2]  $
        \Comment{Pick first K/2 data points}
        \State $\mathcal{\tilde{D}}_{n} \coloneqq \mathcal{D}_{n}[0:K/2]  $
        \Comment{Pick first K/2 data points}
        \State $\mathcal{\tilde{D}} \coloneqq \mathcal{\tilde{D}}_{p} \cup \mathcal{\tilde{D}}_{n} $
        \Comment Merge both sampled data set
        \State \Return $\mathcal{\tilde{D}}$
        \EndFunction
        
        \end{algorithmic}
        \end{algorithm}

    \end{enumerate}

In relational domains, typically, there could be multiple groundings of the same predicate. For instance, {\em ActedIn(P,M)} could yield multiple movies $M$ for the same actor $P$. Therefore, we constructed a contextual bandit problem for the relational classification data in the following way: a regret is counted as 0 (else 1) if and only if the algorithm predicts one of the correct label (i.e., similar to the multi-label classification) of the given data instance correctly. A similar framework in propositional domains is widely adapted in the literature~\cite{cb_backoff,Agarwal,tree_cb}.

\subsection{Experimental Results}

\textbf{[Q1]} How does \textsl{RB2} perform against benchmark algorithm on real-world data sets?

Figure~\ref{results} presents the cumulative regret of \textsl{RB2} with other algorithms on synthetic and real-world data sets. The hyperparameter description that has been employed in our experiments are present in Table~\ref{table2}. In all cases, except the Drug interaction data set, \textsl{RB2} achieves an equal performance or better than all other baselines. Note that in drug interaction, even batch RRTGB without any exploration outperforms all others. One of the caveats is that the data set is relatively easy to learn with only a few initial batches of data forming a representative sample of the whole data set. Therefore, very little exploration is required for an initial set of weak learners to find an optimal solution. Greedy~\textsl{RB2} also shows similar performance on NELL and IMDB data sets. Because after enough training data, \textsl{RB2} eventually learned the data distribution. Thus it does not require the stochasticity needed for exploration in samples. Except for the drug interaction data set, batch RRTGB without exploration has converged to a lower asymptote. Thus the batch learning method failed to learn the accurate context-reward distribution. Our analysis showed that this happens due to overfitting on a given batch sample. The lack of exploration to find the truth labels could be the key reason for overfitting.

\textbf{[Q2]} How does \rb perform against linear approximation model?

Next, we compare our \rb framework against the classic propositional contextual bandit benchmark \textsl{LinUCB}~{L.Li} on IMDB data set. Note that this is just a representative data set and chosen to highlight why relational models are necessary. This experiment's underlying motivation is to evaluate the RRTGB framework's performance against the linear approximation model in the relational domain. \textsl{LinUCB} fits the linear regression model on context-reward relationship for each action. We choose the action with the highest upper confidence bound among all actions with respect to given new context's calculated probability of getting $r=1$.

First, we convert the relational data into the flat-vector representations. We use binary vector representations to encode relations into a feature vector. The resulting feature vector will be very sparse due to the polynomial numbers of possibility of relations. For example, in \textsl{Friends(A,B)} relation with a total of $n$ people, $\binom{n}{2}$ combinations are possible, which is approx. $O(n^2)$. For all the entities and relationships in the domain, we encode into binary vector and perform \textsl{LinUCB} on it.

\begin{figure}[ht]
\centering
\includegraphics[width=\columnwidth,scale=0.2]{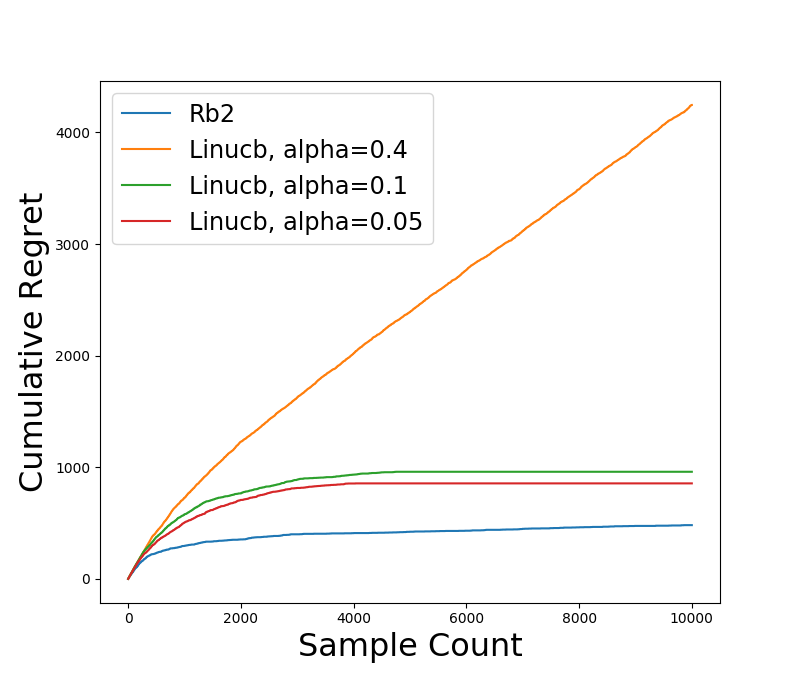} 
\captionof{figure}{Comparison of \rb with \textsl{LinUCB} on IMDB data set. The performance of \textsl{LinUCB} is measured on different values of hyper parameter $\alpha = 0.05,0.1,0.4$ }
\label{fig2}
\end{figure}

Figure \ref{fig2} shows the cumulative regret of \rb and \textsl{LinUCB} on IMDB data set. The $\alpha$ parameter governs the exploration-exploitation trade-off in \textsl{LinUCB}. \rb clearly outperforms the \textsl{LinUCB}. This also shows the importance of using tree learners for binary response data. While the necessity of relational learning methods is well established in literature~\cite{book_srl}, it is necessary to demonstrate this effectiveness in the context of bandits. 

\textbf{[Q3]} How effective is \textsl{RB2} in terms of interpretability?

\begin{figure}[ht]
\centering
\includegraphics[width=\columnwidth]{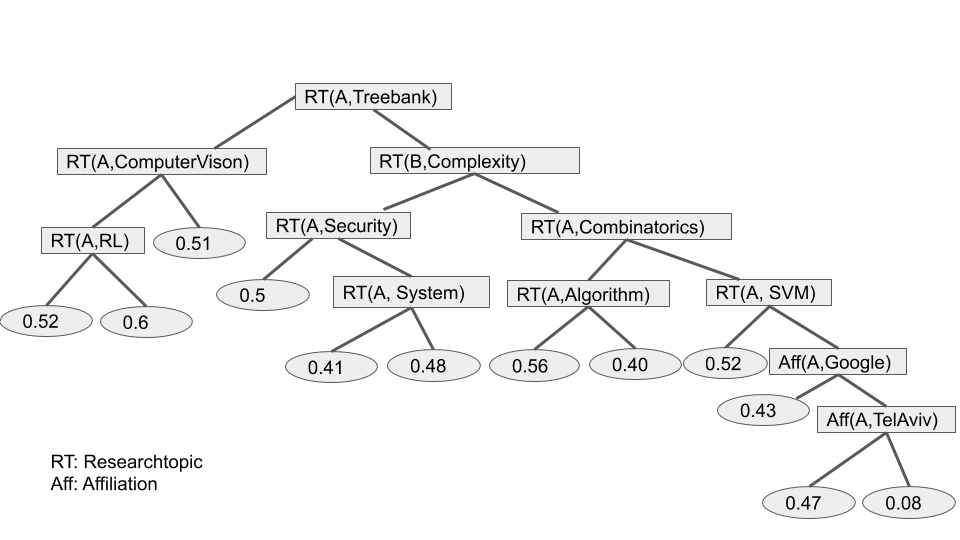} 
\captionof{figure}{Example of a reward prediction model learnt by \rb on ICML dataset. The target here is Co-Author(A,B) }
\label{fig4}
\end{figure}

Figure \ref{fig4} represents the estimated context-reward distribution on the ICML dataset, learned as a single relational regression tree. This tree represents the sum of all the RRTs. The nodes except at the leaf represent the predicates and conjunctions of literals. The leaf nodes represent the probability values of receiving reward $r=1$ for predicting whether author A and B have worked together. If the predicate in an interior node is true, then we traverse the left path, otherwise the right path.

\section{Conclusion}
We presented \rb, a novel contextual bandit algorithm for online learning in relational domains. We use gradient-boosted relational trees as a base learner and softmax exploration for the exploration-exploitation trade-off. We also proposed a parametric-free sampling algorithm that is suitable for online relational learning. We empirically showed the performance of \rb with other benchmark algorithms on the cumulative regret evaluation metric and presented an interpretable tree for evaluation. Considering other exploration techniques for efficient exploration can result in efficient learning could be an interesting direction. Combining active learning strategies with efficient exploration strategies can result in a powerful human-in-the-loop system for online learning. To achieve true human-in-the-loop learning, it is essential that the learned models are explained. Going beyond the empirical combination and constructing an analytical additive combination is an interesting direction for future research.

\section{Acknowledgements}
AK acknowledges the support of Verisk AI Faculty Research Award to BR. SN gratefully acknowledges the support of AFOSR award FA9550-18-1-0462 and ARO award W911NF2010224. Any opinions, findings and conclusions or recommendations expressed in this material are those of the authors and do not necessarily reflect the view of the AFOSR, ARO or the US government.
\section*{Ethics Statement}

Our goal is to develop a new online learning algorithm for learning from noisy, structured data. The major impact that our work aims to create is developing a bandit learner for multi-relational data. Our work has the potential to be employed in large-scale relational classification tasks such as entity resolution and relation extraction. Specifically, we anticipate the use of such algorithms in the context of recommendation systems. Rigorous evaluation on larger data can enable this adaptation.

As far as we are aware, there is no major ethical impact of this work. The only minor negative impacts could be the misinterpretation of the rules by experts but this is a standard issue with any rule learning method. Beyond this aspect, we do not anticipate any major negative societal impact.

\bibliography{rb2.bib}
\end{document}